\documentclass{article}
\usepackage{arxiv}
\usepackage{cite}
\usepackage[utf8]{inputenc} % allow utf-8 input
\usepackage[T1]{fontenc}    % use 8-bit T1 fonts
\usepackage{hyperref}       % hyperlinks
\usepackage{url}  
\usepackage{graphicx}
\usepackage{xcolor}
\usepackage{float}
\usepackage{booktabs}       % professional-quality tables
\usepackage{amsfonts}       % blackboard math symbols
\usepackage{nicefrac}       % compact symbols for 1/2, etc.
\usepackage{microtype}      % microtypography
\usepackage{lipsum}
\usepackage{caption}
\usepackage{authblk}
\usepackage[bottom]{footmisc}
\usepackage{adjustbox}
\usepackage{xcolor}

\title{Multi-class Generative Adversarial Networks for Semi-supervised Image Classification}

\author[1,2,*]{Saman Motamed}
\author[1,2,3]{Farzad Khalvati}
\affil[1]{Institute of Medical Science, University of Toronto}
\affil[2]{Department of Diagnostic Imaging, Neurosciences and Mental Health, The Hospital for Sick Children}
\affil[3]{Department of Mechanical and Industrial Engineering, University of Toronto}
\affil[*]{sam.motamed@mail.utoronto.ca}

\begin{document}
\maketitle
\begin{abstract}
From generating never-before-seen images to domain adaptation, applications of Generative Adversarial Networks (GANs) spread wide in the domain of vision and graphics problems. With the remarkable ability of GANs in learning the distribution and generating images of a particular class, they can be used for semi-supervised classification tasks. However, the challenge is that if two classes of images share similar characteristics, the GAN might learn to generalize and hinder the classification of the two classes. In this paper, we use various images from MNIST and Fashion-MNIST datasets to illustrate how similar images cause the GAN to generalize, leading to the poor classification of images. We propose a modification to the traditional training of GANs that allows for improved multi-class classification in similar classes of images in a semi-supervised learning framework.  
\end{abstract}

% keywords can be removed
\section{Introduction}
Generative Adversarial Networks \cite{goodfellow2014generative} is one of the most exciting inventions in machine learning in the past decade. While applications of GANs spread wide in the field of computer vision, image classification using GANs is relatively unexplored. One of the early uses of GANs in image classification was detecting anomalies in images, first introduced by Schlegl \emph{et al.}~\cite{schlegl2017unsupervised} to detect and identify anomalies in the form of retinal fluid or hyper-reflective foci in optical coherence tomography (OCT) images of the retina. By defining a variation score \(V(x)\) (eq. \ref{eq:ax}), their proposed Anomaly Detection GAN (AnoGAN) captured the characteristic and visual differences of two images; one generated by the GAN and one real image. The idea was to, for instance, train the GAN on only healthy images. When GAN is trained, the generator can generate images similar to those in the healthy image class. During the test phase, the variation score \(V(x)\) must be low if the test image is healthy and GAN's generator (\textbf{G}) can generate a similar image to that of the healthy image. If the test image is not healthy and contains anomalies, \(V(x)\) would be larger, and the generated image would look visually different than the real image containing anomalies.

Recently, we proposed RANDGAN \cite{motamed2020randgan}, a Generative Adversarial Network for binary classification of COVID-19 negative (healthy and viral pneumonia) and COVID-19 positive chest X-ray images without the need to use any COVID-19 positive images for training the model. By training two GANs, one on normal images and one on pneumonia images, we calculated the final \(V(x)\) for test image \(x\). The GAN trained on normal images would result in lower variation score for normal images and higher scores for pneumonia and COVID-19 images. The GAN trained on pneumonia images would result in low variation score for pneumonia images and higher for normal and COVID-19. Adding the variation scores for image \(x\) from each GAN would result in higher \(V(x)\) for COVID-19 positive images compared to COVID-19 negative (normal and pneumonia) images \cite{motamed2020randgan}. Using a similar approach to RANDGAN \cite{motamed2020randgan}, a semi-supervised multi-class (e.g., 3 classes) image classification can be developed. Having labels for two classes out of 3 classes of images, we used GANs to classify these two classes along with a third \textit{unknown} class for which we do not have labels and hence, do not use any of the unknown images to train our model. The first step of multi-class classification is to distinguish images of the \textit{unknown} class from images of the known classes (e.g., COVID-19 positive from negative images). The second step is to classify the images that fall in the \textit{known} images by using one of the GANs trained on either of the known class of images (e.g. normal or pneumonia). It is to be mentioned that we explore multi-class classification in a setting where we have two known classes (\textbf{C1} and \textbf{C2}) of images and an unknown class (\textbf{C3}) of images. This approach can be extended to multiple known classes and an unknown class of images.

\par We observed that, in some instances, training a GAN on images of a class \textbf{C1} generated not only low variation scores for test images of the same class, but also low scores for test images of class \textbf{C2}, hindering the ability to classify \textbf{C1} vs. \textbf{C2}. We hypothesized the reason to be the ability of the GAN's generator \textbf{G}, being trained on \textbf{C1} images, generalizing to learn and generate images that visually look similar to \textbf{C2} images. In this work, we carried out multiple experiments using different datasets to understand how visually similar images affect GAN-based image classification's performance. We propose \emph{MCGAN}, a GAN-based multi-classs classifier, to overcome the challenge of classifying visually similar images using GANs. By using all labeled images in training the GANs, we force \textbf{G} to not generalize in a way that can generate similar images to images of other classes.

\section{Generative Adversarial Networks}
\subsection{Generative Adversarial Nets}
A GAN is a deep learning model comprised of two main parts; Generator (\textbf{G}) and Discriminator (\textbf{D}). \textbf{G} can be seen as an art forger that tries to reproduce art-work and pass it as the original. \textbf{D}, on the other hand, acts as an art authentication expert that tries to tell apart real from forged art. Successful training of a GAN is a battle between \textbf{G} and \textbf{D} where if successful, \textbf{G} generates realistic images and \textbf{D} is not able to tell the difference between \textbf{G}'s generated images compared to real images.
\textbf{G} takes as input a random Gaussian noise vector and generates images through transposed convolution operations. \textbf{D} is trained to distinguish the real images \((x)\) from generated fake images \((G(z))\). Optimization of \textbf{D} and \textbf{G} can be thought of as the following game of minimax~\cite{goodfellow2014generative} with the value function $V(G, D)$: 
\begin{equation} \label{eq:1}
\min_G \max_D V(D, G) = \mathbb{E}_{x_{{\sim_P}_{data{(x)}}}} [\log D(x)] + \mathbb{E}_{z_{{\sim_P}_{z{(z)}}}} [\log (1 - D(G(z)))]
\end{equation}
During training, \textbf{G} is trained to minimize \textbf{D}'s ability to distinguish between real and generated images, while \textbf{D} is trying to maximize the probability of assigning "real" label to real training images and "fake" label to the generated images from \textbf{G}. The Generator improves at generating more realistic images while Discriminator gets better at correctly identifying between real and generated images. Today, when the term GAN is used, the Deep Convolution GAN (DCGAN) \cite{radford2015unsupervised} is the architecture that it refers to.

\subsection{Multi-class Discriminator GAN}
The goal of the proposed GAN-based multi-class (MCGAN) classifier is to distinguish three classes of data (\textbf{C1}, \textbf{C2}, \textbf{C3}) from one another, while only two classes (\textbf{C1} and \textbf{C2}) have labels and the third class (\textbf{C3}) is unknown and has no labels. In doing so, first, we classify \textbf{C1} and \textbf{C2} vs. \textbf{C3}, and then classify \textbf{C1} vs. \textbf{C2}. Since we have labels for only \textbf{C1} and \textbf{C2}, the first step is to learn the distribution of these two classes using two GANs. Here, we describe the architecture for learning the distribution of \textbf{C1} images and then classifying the images of \textbf{C1} vs. \textbf{C2}. This can be repeated to learn the distribution of \textbf{C2} images and then classifying the images of \textbf{C2} vs. \textbf{C1}. To learn the distribution of \textbf{C1} images, a traditional GAN's (DCGAN, AnoGAN, etc.) discriminator takes as input the generator's output (labeled \textit{Fake}) and a real \textbf{C1} image (labeled \textit{Real}). This forces the generator to learn the distribution of the images from the real class. If the images of \textbf{C1} and \textbf{C2} shares similar characteristics, training the GAN on the images of \textbf{C1} could cause \textbf{G} to learn and generalize well enough, leading to generating similar images to \textbf{C2} and hence, hindering the classification of the two known classes (\textbf{C1} vs. \textbf{C2}). To overcome this challenge, we feed a third input to the discriminator; images of \textbf{C2}.

\begin{figure}[h!]
  \centering
  \includegraphics[scale=0.45]{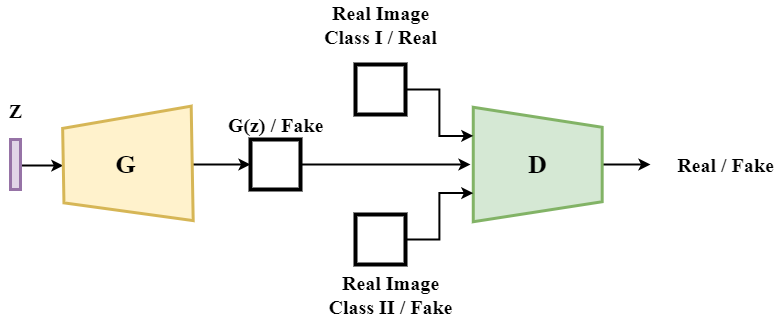}
  \caption{Multi-class Generative Adversarial Network}
  \label{fig:1}
\end{figure}

While these are real images from \textbf{C2}, we label them as \textit{Fake}. This forces the generator not to learn to generalize to this similar class (\textbf{C2}) while learning the characteristics of \textbf{C1}. When \textbf{G} generates an image that could pass as belonging to \textbf{C2}, the discriminator flags it as a fake image, and \textbf{G} re-evaluates its learning at those stances. Figure \ref{fig:1} shows the architecture of Multi-class GAN (MCGAN). In imbalanced datasets where the number of images is not the same for both C1 and C2, we use batches of randomly selected images of C2 at each training iteration. 

\subsection{Variation Score}
The Variation score $V(x)$ for the query image $x$, proposed by Schlegl \emph{et al.} \cite{schlegl2017unsupervised}, is defined as;
\begin{equation} \label{eq:ax}
V(x) = (1 - \lambda) \times \mathcal{L}_{R}({z}) + \lambda \times \mathcal{L}_{D}({z})
\end{equation}
where \(\mathcal{L}_{R}({z})\) (eq.~\ref{resi}) and \(\mathcal{L}_{D}({z})\) (eq.~\ref{disc}) are the residual  and discriminator loss respectively that enforce visual and image characteristic similarity between real image x and generated image \(G(z)\). The discriminator loss captures image characteristics using the output of an intermediate layer of the discriminator, \(f(.)\), making the discriminator act as an image encoder. Residual loss is the pixel-wise difference between image x and \(G(z\)).
\begin{equation}\label{resi}
\mathcal{L}_R({z}) = \sum|x - G(z)|
\end{equation}
\begin{equation}\label{disc}
\mathcal{L}_D({z}) = \sum|f(x) - f(G(z)|
\end{equation}
Before calculating V(x) in test, a point $z_i$ has to be found through back-propagation that tries to generate an image as similar as possible to image x. The loss function used to find $z_i$ is based on residual and discriminator loss defined below.
\begin{equation}\label{eq:4}
  \mathcal{L}({z_i}) = (1 - \lambda) \times \mathcal{L}_{R}({z_i}) + \lambda \times \mathcal{L}_{D}({z_i})
\end{equation}
$\lambda$ adjusts the weighted sum of the overall loss and variation score. We used $\lambda = 0.2$ to train our proposed MCGAN and AnoGAN~\cite{schlegl2017unsupervised}. Both architectures were trained with the same initial conditions for performance comparison.

\section{Datasets}
We used images from two different datasets. MNIST \cite{lecun-mnisthandwrittendigit-2010} dataset that contains 60,000 training images of handwritten digits and 10,000 test images. Fashion-MNIST \cite{xiao2017fashion} is a dataset of Zalando's article images—consisting of a training set of 60,000 examples and a test set of 10,000 examples. In experiments where test sets are not balanced, we randomly select the same number of images as the smaller test set for class \emph{a} from the bigger test set of class \emph{b}.
All gray-scale images were resized to \(64 \times 64\) pixels, with pixel intensities scaled to -1 to 1.

\begin{figure}[h!]
  \centering
  \includegraphics[scale=0.3]{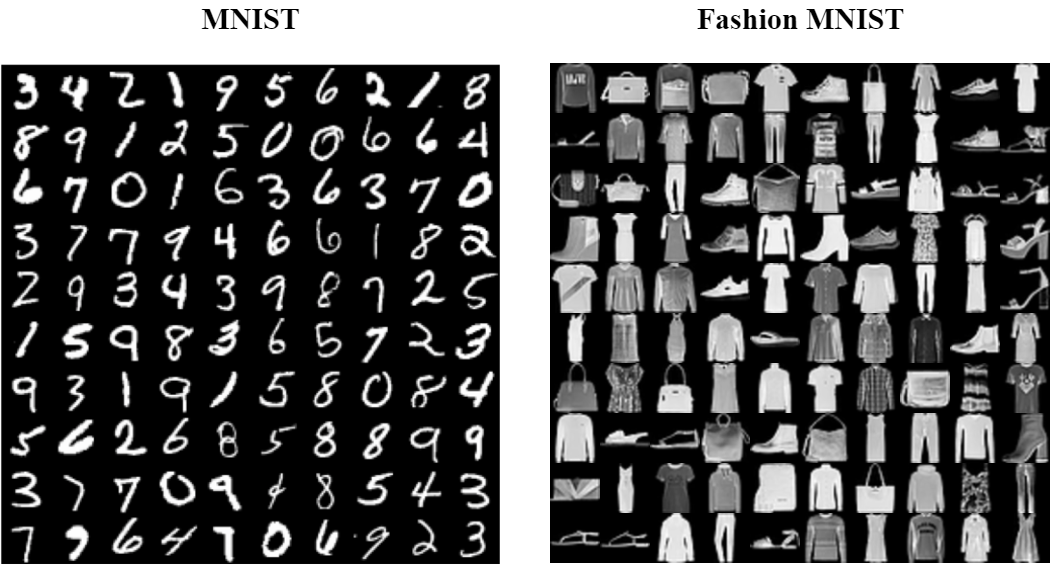}
  \caption{MNIST and Fashion MNIST sample images}
  \label{fig:DATA}
\end{figure}

\section{Experiments}
\begin{figure}[h!]
  \centering
  \includegraphics[scale=0.35]{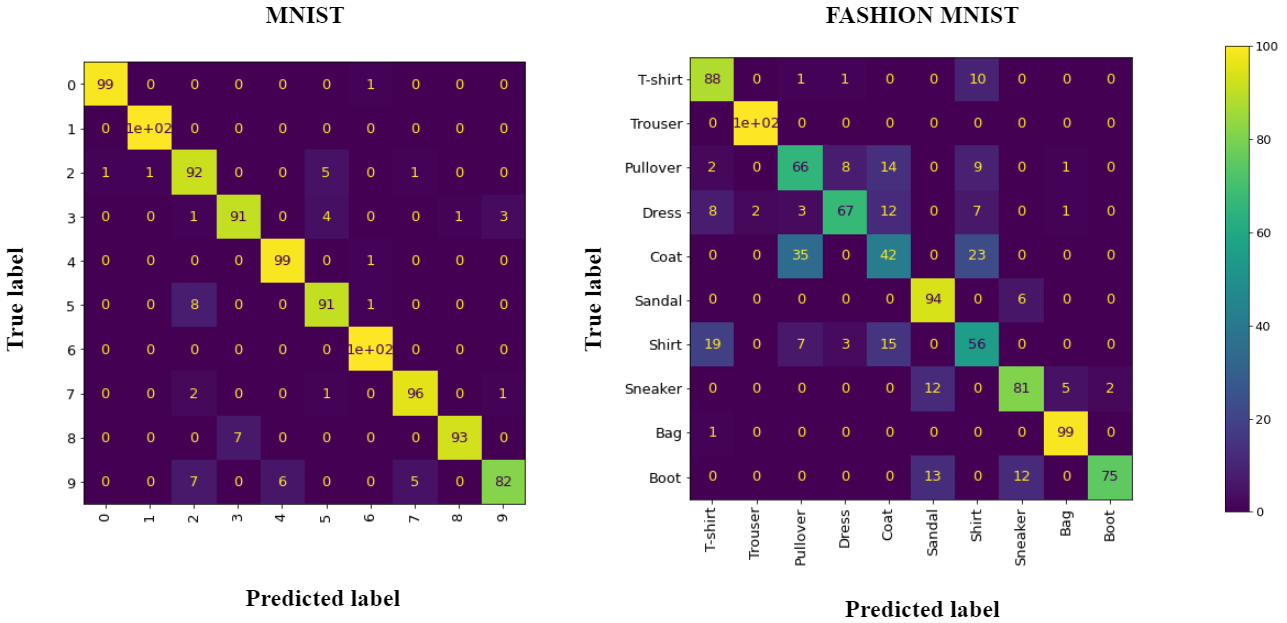}
  \caption{Confusion matrix of MNIST and F-MNIST embeddings}
  \label{fig:2}
\end{figure}

To pick a subset of similar classes from MNIST and Fashion-MNIST (F-MNIST) datasets that could cause generalization in GANs, we used metric learning \cite{kulis2012metric}. The goal of metric learning is to train models that can embed inputs into a high-dimensional space such that "similar" inputs are located close to each other. To bring images from the same class closer to each other via the embedding, the training data was constructed as randomly selected pairs of images from each class matched to the label of that class, instead of traditional \textit{(X,y)} pairs where y is the label for corresponding X as singular images of each class. By embedding the images using a shallow three-layer CNN, we computed the similarity between the image pairs by calculating the cosine similarity of the embeddings. We used these similarities as logits for a softmax. This moves the pairs of images from the same class closer together. After the training was complete, we sampled 10 examples from each of the 10 classes, and considered their near neighbours as a form of prediction; that is, does the example and its near neighbours share the same class. This is visualized as a confusion matrix shown in figure \ref{fig:2}. The numbers that lie on the diagonal represent the correct classifications and the numbers off the diagonal represent the wrong labels that were misclassified as the true label. We intentionally used a shallow three-layer CNN to enforce some misclassification, as achieving near-perfect results in classifying datasets such as MNIST using CNNs is easy. Using the information from figure \ref{fig:2}, we picked the class pairs (9, 4) and (8, 3) from the MNIST dataset and (Coat, Shirt), (Coat, Pullover), and (Boot, Sandal) from F-MNIST dataset.

\par For semi-supervised multi-class classification of the pair of \textit{known} images that we have labels for and an \textit{unknown} class which we will introduce when testing our models, we trained two GANs. One GAN was trained to generate images similar to each class of known images. For instance, for the pair (9, 4), one DCGAN (AnoGAN and DCGAN have the same architecture) was trained on 9s and one was trained on 4s. Similarly, one MCGAN was trained on 9s, labeled \textit{Real} while 4s were labeled \textit{Fake} and one MCGAN was trained on 9s, labeled \textit{Fake} while 4s were labeled \textit{Real}. The models were trained using an NVIDIA GeForce RTX 2080 Ti with 11 GB of memory. 

\section{Results}
For classification of images from known C1 and C2 and unknown C3 classes, first we classify the unknown images from known images by looking at the final variation score which is the sum of variation scores from each trained GAN (the one GAN trained on C1 and the one trained on C2). After classifying C3 images, we need to classify C1 and C2 images. For this, we have two options; 1) use the GAN trained on C1 images to classify the remaining images as C1 or C2 or 2) use the GAN trained on C2 images and perform classification. In order to pick the best instance for step 2, we look at the performance of supervised binary classification of the known classes C1 and C2. The GAN that performs better in classifying C1 and C2 is used for the second step of multi-class classification in the semi-supervised setting when we introduce images of class C3. Figure \ref{fig:4} shows this process.

\subsection{Supervised binary classification}
We calculated variation scores for both DCGANs and MCGANs. Lower variation scores would translate to the test image having more probability of belonging to the class of images the GAN was trained to generate images of, while a larger variation score decreased this probability. We calculated the area under the ROC curve (AUC) of each model. Table \ref{auc} shows the AUC for binary classification for each data pair and model. For each pair (C1, C2) for DCGAN, the first AUC is the result of training GAN on C1 images and the second AUC is the result of GAN training on C2 images. For MCGAN, the first AUC is the result of using C1 with \textit{Real} labels and C2 with \textit{Fake}, while the second AUC is for vise versa. In both scenarios of using MCGAN to generate images of class C1 or C2, MCGAN outperformed DCGAN in binary classification.
\begin{table}[h!]
\begin{adjustbox}{width=\columnwidth,center}
 \begin{tabular}{||c c c c c c ||} 
 \hline
 & MNIST & MNIST & F-MNIST & F-MNIST & F-MNIST\\
 & (8 / 3) & (9 / 4) & (Boot / Sandal)  & (Coat / Shirt) & (Coat / Pullover)\\ [0.5ex] 
 \hline\hline
 DCGAN & 0.87 / 0.9 & 0.88 / 0.78 &  0.65 / 0.72 & 0.68 / 0.54 & 0.67 / 0.32\\
 \hline
 MCGAN & \textbf{0.92 / 0.95} & \textbf{0.91 / 0.84} &  \textbf{0.87 / 0.82} & \textbf{0.79 / 0.74}& \textbf{0.77 / 0.71} \\
 \hline
\end{tabular}
\end{adjustbox}
\caption{Classification AUCs}
\label{auc}
\end{table}

\subsection{Semi-supervised multi-class classification}
\begin{figure}[h!]
  \centering
  \includegraphics[scale=0.25]{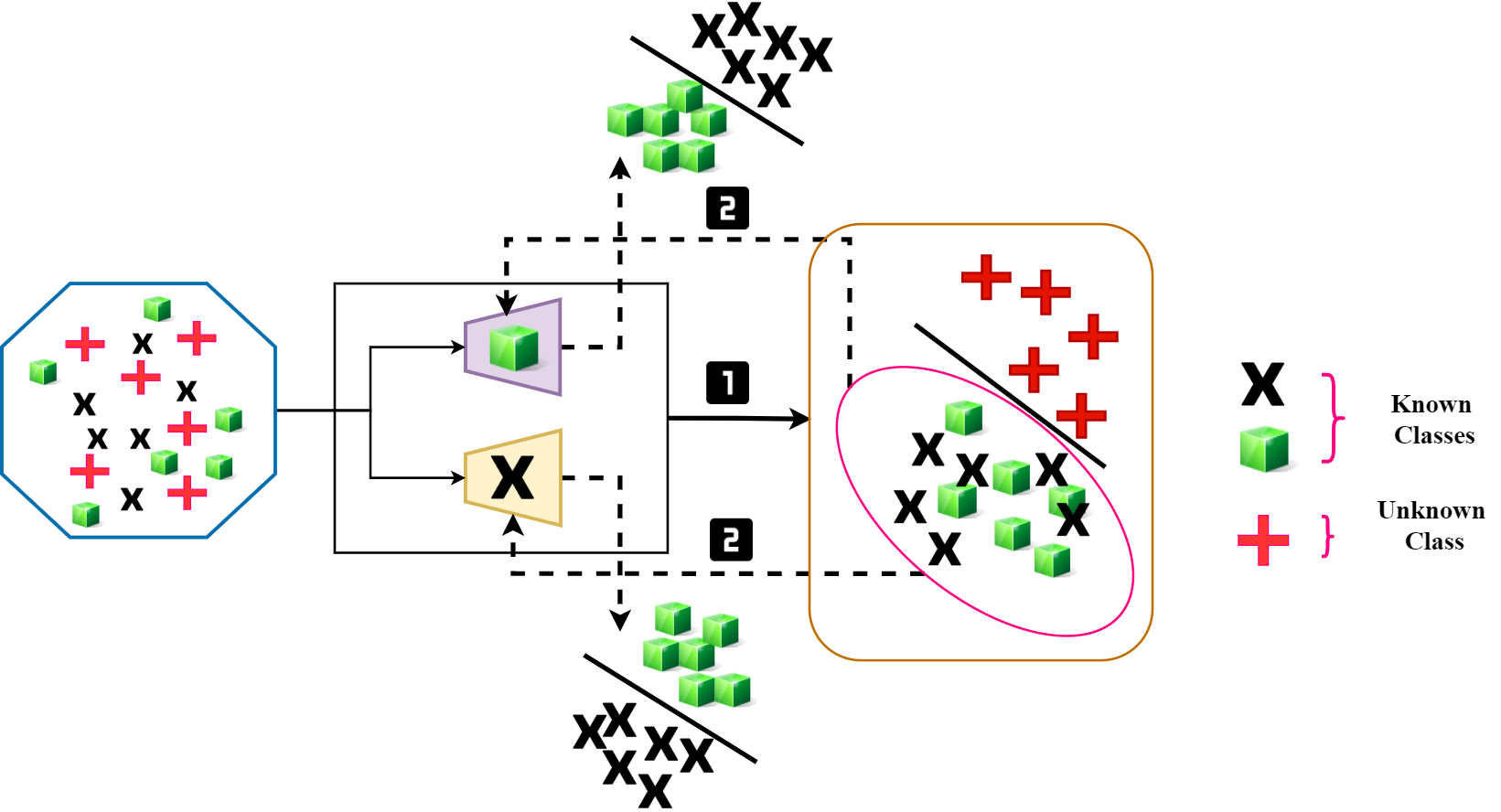}
  \caption{Two step process of multi-class classification using two GANs trained on known classes}
  \label{fig:4}
\end{figure}

For semi-supervised multi-class classification, we added images from class 9 to (3, 8), 8 to (4, 9), Dress to (Coat, Pullover), Bag to (Boot, Sandal) and Pullover to (Coat / Shirt). First, we calculate the accuracy of classifying the unknown class from known classes, then, by using results from table \ref{auc}, we picked the GAN that performed better in classifying the known pairs, and used that GAN to label the remaining test images that were not categorized as the unknown class - C3. Figure \ref{fig:4} shows the two step process of multi-class classification using the two trained GANs. In step 1, the combined variation scores from the two trained GANs on known classes scores the unknown class images higher than known classes. This allows for separation of known classes from the unknown class. In the second step, we can use either GAN (we picked the one with better performance according to table \ref{auc}) to classify the known classes from one another.

Table \ref{aucsemi}, with the format (C1, C2 - C3), shows the accuracy of classifying C3 (unknown class) from C1 and C2 (known classes used in training) and C1 from C2. Number \textbf{a} from the pair (a / b) in the table shows the accuracy of classifying the unknown class C3 and \textbf{b} shows the accuracy of classifying C1 from C2. MCGAN outperformed multi-class classification compared to DCGAN in all but one case where DCGAN's classification accuracy of Bag from Sandal and Boot was 80\% compared to MCGAN's 79\%. For the classification of known classes however, MCGAN outperformed DCGAN in all instances.

\begin{table}[h!]
\begin{adjustbox}{width=\columnwidth,center}
 \begin{tabular}{||c c c c c c||} 
 \hline
 & MNIST & MNIST & F-MNIST & F-MNIST & F-MNIST\\
 & (3, 8 - 9) & (9, 4 - 8) & (Sandal, Boot - Bag) & (Coat, Shirt - Pullover) & (Coat, Pullover - Dress)\\ [0.5ex] 
 \hline\hline
 DCGAN & 0.67 / 0.57 & 0.87 / 0.79 & \textbf{0.8} / 0.77 & 0.66 / 0.42 & 0.73 / 0.52 \\
 \hline
 MCGAN & \textbf{0.69 / 0.64} & \textbf{0.9 / 0.84} & 0.79 / \textbf{0.81} & \textbf{0.67 / 0.65} & \textbf{0.79 / 0.65}\\
 \hline
\end{tabular}
\end{adjustbox}
\caption{Semi-supervised classification AUCs}
\label{aucsemi}
\end{table}

\begin{figure}[h!]
  \centering
  \includegraphics[scale=0.3]{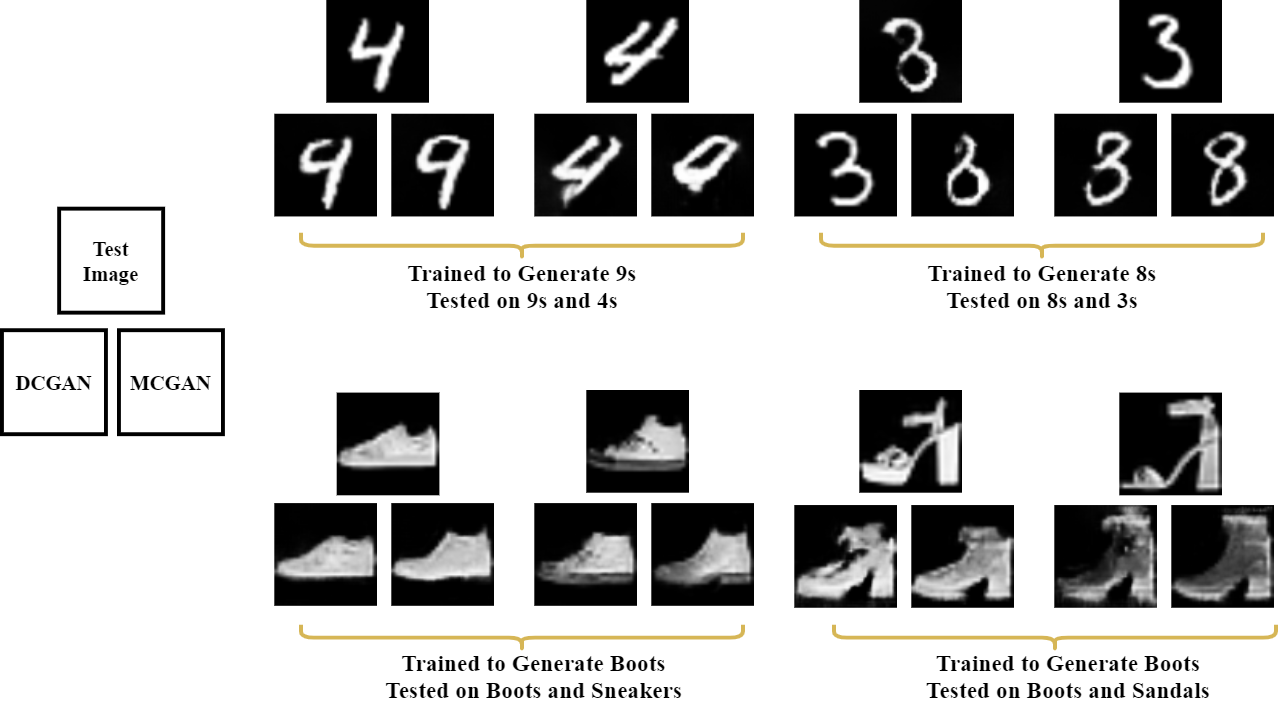}
  \caption{DCGAN and MCGAN generated images}
  \label{fig:3}
\end{figure}

    Figure \ref{fig:3} shows how a DCGAN, trained on images from C1, can generate images similar to C2 if both classes have similar characteristics in a way that while learning to generate C1 images, \textbf{G} learns to produce similar images to C2. MCGAN, on the other hand, forces the generator to avoid this generalization, which helps the classification task (table \ref{auc}). The first row of figure \ref{fig:3} shows the output of DCGAN and MCGAN's generators trying to generate a similar image to the test image from class \textbf{4} and \textbf{3} while having been trained to generate images of class \textbf{9} and \textbf{8} respectively. The DCGAN's output looks closer to the test image compared to MCGAN. The second row result from the two GANs trained to generate images of class \textbf{Boot} and generate images similar to test images from class \textbf{Sneaker} and \textbf{Sandal} respectively. While DCGAN learns to generate images similar to Sneaker and Sandal while learning from images from class Boot, our MCGAN succeeds in not making this generalization. 
For the pair Coat and Shirt, we picked the Pullover class as the unknown class on purpose as all three classes are suggested to be similar according to figure \ref{fig:2}.

\par The semi-supervised classification using MCGAN outperformed DCGAN in classifying the unknown class in all but one experiment (Boot, Sandal - Bag) where the accuracy of detecting the unknown class (bag) reduced from 80\% to 79\% using MCGAN while the classification of the known classes improved from 77\% to 81\%. MCGAN enabled a better classification of the known classes (Tables \ref{auc}, and\ref{aucsemi}) while improving the classification of the unknown class in all but one experiment (Table \ref{aucsemi}).

\section{Discussion}
In classification settings where we do not have enough labeled images for a class, semi-supervised modes of training that do not require images of that class to train are of value. While GANs can be used to classify images, we showed that in some settings where labeled images share similar characteristics, the generalization ability of GANs can hinder the performance of classification. Using images from MNIST and Fashion MNIST datasets, we showed how, for instance, a GAN trained to generate images of handwritten digit 8 can also generate images that are similar to digit 3. To use GANs in classifying 8s from 3s, this generalization would result in not only low variation scores for images of digit 8, but also for images of digit 3. We proposed MCGAN, which used both classes in training the GAN's discriminator. By labeling the digits 3 as \textit{fake}, we guided the generator to not generate images that can identify as 3 while learning to generate images of class 8. This improved the multi-class classification of both the unknown class from known classes and the known classes from one another.
MCGAN however, as shown in table \ref{aucsemi} with classes (Coat, Shirt - Pullover) and (3, 8 - 9), although works better than DCGAN, still does not perform well in the task of multi-class classification. When the unknown class (Pullover, 9) shares similar characteristics with some / all of the known classes, while MCGAN shows improved results in classifying the classes compared to DCGAN, the classification performance suffers as a result. 

\par The goal for this study was not to achieve state of the art classification results on the two datasets, rather using a simple GAN architecture and showing how the proposed modification in training the discriminator can improve classification in settings where over-generalization is possible. With development of more complicated GAN architectures, such as RANDGAN \cite{motamed2020randgan} for detection of COVID-19 X-rays, this modification can further improve the accuracy of the models.

\section{Conclusion}
In this work, we demonstrated how GANs could learn to generalize to different classes of images if they share similar characteristics with the class of training images. This generalization can hinder the ability of GANs for the task of image classification. We proposed using all labeled images in training the discriminator to penalize the generalization. The multi-class discriminator training showed improved accuracy of semi-supervised image classification.
\section{Acknowledgements}
This research was funded by Chair in Medical Imaging and Artificial Intelligence funding, a joint Hospital-University Chair between the University of Toronto, The Hospital for Sick Children, and the SickKids Foundation.
\bibliography{references}

\end{document}